\documentclass[a4paper]{article}

\usepackage{INTERSPEECH2021}
\usepackage{url}
\usepackage{microtype}
\usepackage{multirow}
\usepackage{multicol}
\usepackage{array}
\usepackage{threeparttable}
\usepackage{svg}
\usepackage{siunitx}
\title{Token-Level Supervised Contrastive Learning for Punctuation Restoration}
\name{{Qiushi Huang$^{1,2}$, Tom Ko$^{1*}$, H Lilian Tang$^2$, Xubo Liu$^{2}$, Bo Wu$^3$} \thanks{* corresponding author}}
\address{
  $^1$Department of Computer Science and Engineering\\ Southern University of Science and Technology, Shenzhen, China\\
  $^2$School of Computer Science and Electronic Engineering, University of Surrey, UK\\
  $^3$MIT-IBM Watson AI Lab, Cambridge, USA
}
\email{qiushi.huang@surrey.ac.uk, tomkocse@gmail.com,\{h.tang, xubo.liu\}@surrey.ac.uk, bo.wu@ibm.com}

\begin{document}

\maketitle
\begin{abstract}
Punctuation is critical in understanding natural language text. Currently, most automatic speech recognition (ASR) systems do not generate punctuation, which affects the performance of downstream tasks, such as intent detection and slot filling. This gives rise to the need for punctuation restoration. Recent work in punctuation restoration heavily utilizes pre-trained language models without considering data imbalance when predicting punctuation classes. In this work, we address this problem by proposing a token-level supervised contrastive learning method that aims at maximizing the distance of representation of different punctuation marks in the embedding space. The result shows that training with token-level supervised contrastive learning obtains up to $3.2\%$ absolute $F_1$ improvement on the test set.\footnote{The code is available at \url{https://github.com/hqsiswiliam/punctuation-restoration-scl} }
\end{abstract}

\section{Introduction}\label{sec:introduction}

Punctuation symbols are often absent in the transcript generated by the automatic speech recognition (ASR) system. That often causes degradation in readability for both humans and machines \cite{RNN3}. Removing punctuation from transcriptions notably impacts the comprehension of text. For subsequent machine learning tasks, such as intent detection or slot filling, the performance suffers from missing punctuation since such models usually are trained on clean text with punctuation marks. Thus, it is essential to predict and insert punctuation for the speech transcripts.

Many methods have been proposed to tackle this problem. They can generally divide into three paradigms. The first approach considers the prosodic feature as an essential clue to punctuation insertion \cite{prosody1, prosody2}. It is natural to consider adding punctuation by the prosody of the speech since humans often adopt this approach to add punctuation in their minds. Nevertheless, the acoustic prosody feature is often noisy and error-prone, leading to a sub-optimal result. The second paradigm fuses prosodic features with lexical information. These methods incorporate acoustic and textual features by embedding different features \cite{TRANS_NOT_BERT,Proso_Lex_approach1,Proso_Lex_approach2}. However, the dataset with the acoustic and textual features is not always readily available. The third approach is to incorporate just the lexical information in the transcripts. Early attempts of lexical approach on restoring punctuation utilize n-gram language model that treats the prediction as hidden events \cite{ngram1,ngram2,ngram3}. Later, some proposed work demonstrates the methods based on recurrent neural network (RNN) on the punctuating task \cite{RNN1,RNN2,RNN3}. Unlike previous attempts, RNN can be better capture textual context within the segment than the n-gram methods. Recently, the transformer \cite{attention} based language model dominates language-related tasks, which has better performance and the capability to capture contextual information across longer distances than the RNN. 
Meanwhile, several transformer-based language models \cite{BERT,RoBERTa,albert} that are pre-trained on gigabytes of corpora greatly enhance the performance of downstream tasks, such as Named Entity Recognition. A few approaches through pre-trained models for the punctuating task have been proposed with promising performance \cite{2020_Disflu_BERT,focal_loss_bert,acl2020-punc,BERT_Latest,BERT1}.
This work will explore this approach as our based model and work with textual data only.
Therefore, the problem can be as well simplified to the textual sequential labeling task.

As most existing works ~\cite{2020_Disflu_BERT,focal_loss_bert,acl2020-punc,BERT_Latest,BERT1} consider this problem a token-level classification task, it leads to a data imbalance problem.
In the IWSLT2011\footnote{\url{http://hltc.cs.ust.hk/iwslt/index.php/evaluation-campaign/ted-task.html}} dataset, which is commonly used in the automatic punctuation restoration task, 
over $85\%$ of tokens are NO PUNCTUATION, and only less than $15\%$ of them are with punctuation labels.
Moreover, the question mark accounts for only $0.54\%$ of the total data, much less than that of other classes. The class imbalance problem will influence the performance of the punctuating task and should be addressed.

Contrastive learning \cite{CPC, supervised_contrastive_learning} can leverage the information from all classes rather than those from the corresponding class during training. By contrasting information from all labels, the problem of data imbalance can be alleviated. This paper incorporates supervised contrastive learning (SCL) to demonstrate the learning effectiveness from imbalanced data. Unlike the previous method, which tries to solve this problem \cite{focal_loss_bert} by adopting the weighted term to loss, contrastive learning contrasts the information from the same class against other classes utilizing all the information within a batch in the loss calculation. Meanwhile, clusters of the same classes in the embedding space are pulled together, making the classifier trivial to find the boundaries in the latent space. One linear layer that follows the transformer-based model can effectively classify the punctuation without complex structures like bidirectional LSTM \cite{crf_transformer1, BERT_LSTM_CRF} or transformers \cite{ct_transformer}.

The main contribution of this paper is to incorporate token-level supervised contrastive learning to address the data imbalance problem in punctuating restoration. Experiments conducted on the IWSLT dataset show the models trained with this approach gain up to $3.2\%$ absolute overall $F_1$ score than the model trained with just the cross-entropy loss.

The rest of the paper is organized as follows: Section \ref{sec:review} reviews the contrastive learning theory. 
Section \ref{sec:method} describes our proposed method, followed by detailed experiments in
Section \ref{sec:exp}. 
Section \ref{sec:conlusion} concludes and opens up points for future work.

\begin{figure}[t]
  \centering
  \includegraphics[width=\linewidth]{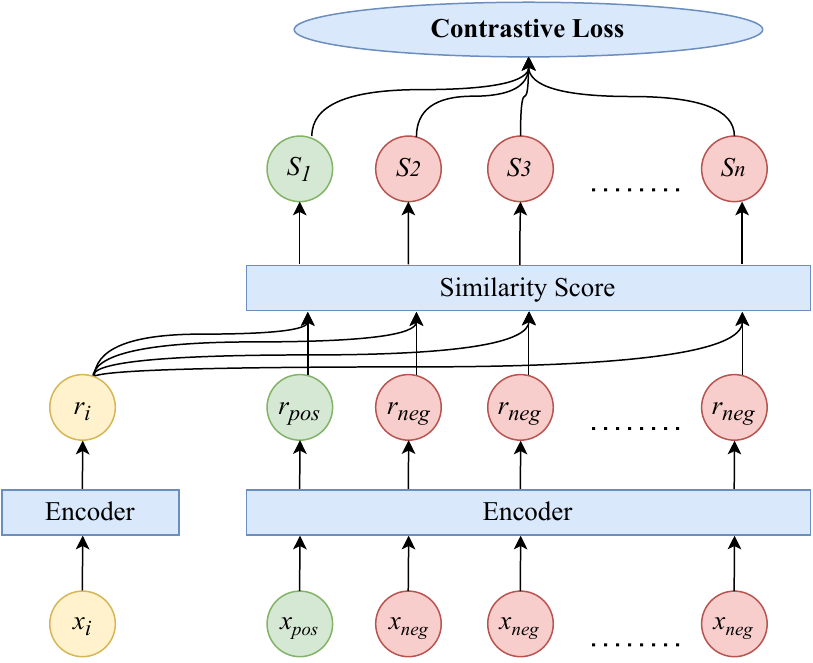}
  \caption{Given an anchor sample $x_i$, a positive sample $x_{pos}$, and negative samples $\{x_{neg}\}$ in a batch, the encoder encodes each sample into latent space as the representation. The similarity scores $\{S_1,..., S_n\}$ are calculated through the pairs of the anchor and each positive/negative sample. Afterward, a contrastive loss is calculated based on the similarity scores.}
  \label{fig:contrastive_learning}
  \vspace{-4mm}
\end{figure}

\section{Preliminaries}\label{sec:review}
\subsection{Contrastive learning}
The concept of contrastive learning ~\cite{CPC} builds on self-supervised learning (SSL). Self-supervised learning is a learning paradigm to capture the inherent patterns and context in data without human labeling. Therefore, self-supervised learning often constructs pretext tasks solely based on the unlabeled data and forces the network to train with these tasks to learn meaningful representation from data. Contrastive learning is to explore this approach through contrasting samples. As shown in Figure~\ref{fig:contrastive_learning}, given an anchor sample, a positive sample, and negative samples, similarities are calculated pairwise between the given anchor and the rest. A positive sample means it belongs to the same class as the anchor sample (e.g., augmented image of the anchor image or different time slices from audio). In contrast, negative samples mean those not in the same class as the anchor. Then, a contrastive loss is computed using the similarity scores. The formula of the self-supervised contrastive loss takes the following form.

\begin{equation} \label{eq:cl}
\mathcal{L}_{CL}=-\sum_{i \in I} \log \frac{f(z_{i}, z_{j(i)})}{\sum_{k\in A(i)} f(z_k, z_i)}
\end{equation}

Here, $f(z_i,z_{j(i)})=\exp(z_i\cdot z_{j(i)}/\tau )$ calculates the similarity between $z_i$ and $z_{j(i)}$. $\tau$ is the temperature, a scalar to stabilize the calculation. $z_i = \text{Encoder}(x_i)$ denotes the representation calculated by the encoder. $i$ denotes the anchor sample; $j(i)$ denotes its positive sample; $A(i)$ is the set that contains the positive sample and the negative samples.


\subsection{Supervised contrastive learning}
In Supervised Contrastive Learning (SCL) ~\cite{supervised_contrastive_learning}, prior knowledge in the labeled data can be utilized. Since the anchor label is known, the label information can identify the positive and negative samples. 
Therefore, equation~\ref{eq:cl} can be generalized as follows.

\begin{equation} \label{eq:scl}
\mathcal{L}_{SCL}=\sum_{i \in I} \frac{-1}{|P(i)|} \sum_{p\in P(i)} \log \frac{f(z_{i}, z_{p})}{\sum_{k \in A(i)} f(z_k, z_i)}
\end{equation}

The $P(i)$ presented in the $\mathcal{L}_{SCL}$ means all positives correspond to anchor $i$, retrieved by the label information within a batch. The $|P(i)|$ means the number of items in this set.

\begin{figure}[t]
  \centering
  \includegraphics[width=\linewidth]{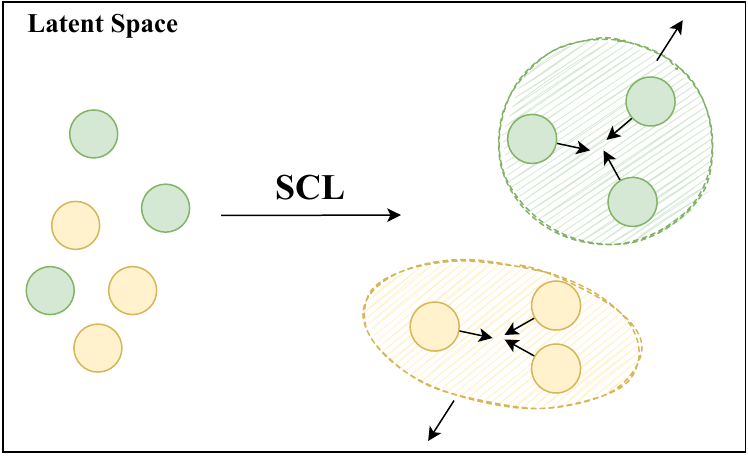}
  \caption{The representations belonging to the same class are pulled together in the latent space while simultaneously pushing apart clusters of samples from different classes by Supervised Contrastive Learning (SCL).}
  \label{fig:pushing away}
  \vspace{-4mm}
\end{figure}

By adopting supervised contrastive learning in training, representations belonging to the same class are pulled together in the latent space while simultaneously pushing apart clusters of samples from different classes, as shown in Figure~\ref{fig:pushing away}. Supervised contrastive learning is closely related to the triplet loss ~\cite{supervised_contrastive_learning}, one of the widely used losses in face recognition ~\cite{facenet}, speaker identification tasks ~\cite{tristounet}. The triplet loss is a special case of supervised contrastive learning where only one negative is used.

\section{Approach}\label{sec:method}
There are two simultaneous objectives during the training. The first one is to train the representations of each token that have distinctive clusters for different labels. The second is to build the boundaries among these clusters to accomplish the classification task. To achieve this, we incorporate a supervised contrastive learning term with the standard cross-entropy loss.

\subsection{Problem setting}
Given an input sequence $X=\{x_1, x_2, ..., x_n\}$ where $n$ denotes the length of the sequence, each $x_i \in X$ denotes the word in a document. The input $X$ is encoded into representation vectors in latent space as $R=\text{Encoder}(X)$ where $R=\{r_1, r_2, ..., r_n\}$. Then, the ground truth is formalized as $Y=\{y_1, y_2, ..., y_n\}$ where length $n$ in $Y$ is equal to the length $n$ in $X$. Meanwhile, the $y_i$ is a predefined list of the four possible punctuation in the documents, which can be \emph{O (No Punctuation), COMMA, PERIOD, QUESTION (Question Mark)}. Afterward, the predicted result from the model is formalized as $\hat{Y}=\{\hat{y_1}, \hat{y_2}, ..., \hat{y_n}\}$.
\begin{figure}[t]
  \centering
  \includegraphics[width=\linewidth]{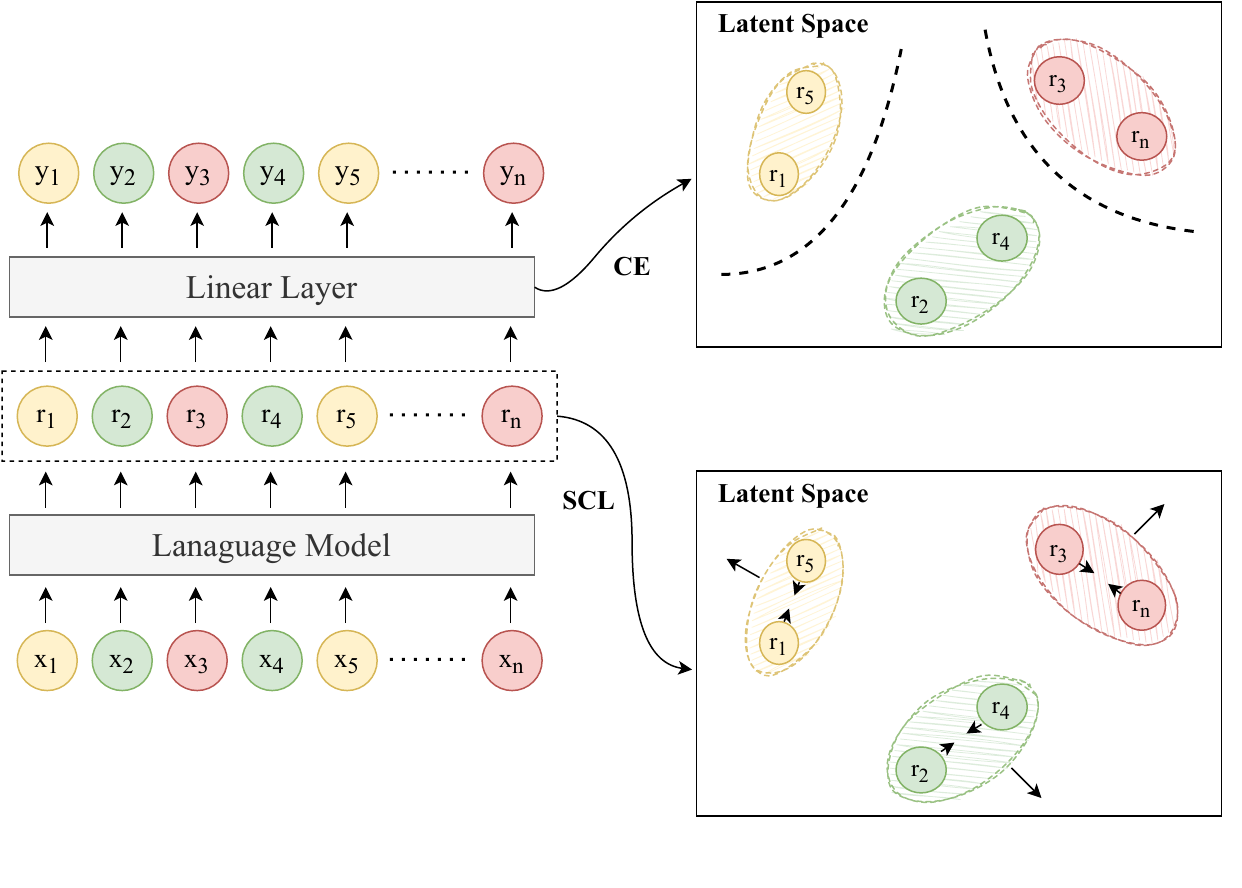}
  \caption{The supervised contrastive learning (SCL) separates the representations of different classes. Meanwhile, cross-entropy (CE) draws the decision boundaries onto the clusters built by SCL.}
  \label{fig:our contrastive learning}
  \vspace{-4mm}
\end{figure}
\subsection{Token-level supervised contrastive learning}


In contrast to the work in ~\cite{Gunel2020SupervisedCL}, which applies SCL to contrast sentences within a batch, we apply token-level SCL to contrast words within a batch. To leverage the label information into both contrastive learning and supervised learning, we add supervised contrastive learning with cross-entropy concurrently, which can be given by the following formula.
\begin{equation}\label{eq:proposed1}
    \mathcal{L}=(1-\lambda)\mathcal{L}_{CE} + \lambda \mathcal{L}_{SCL}
\end{equation}
\begin{equation}\label{eq:proposed2}
    \mathcal{L}_{CE}=-\frac{1}{N} \sum_{i=1}^N \sum_{c=1}^C y_{i,c} \cdot \log \hat{y}_{i,c}
\end{equation}
\begin{equation}\label{eq:proposed3}
\mathcal{L}_{SCL}=\sum_{i \in I} \frac{-1}{|P(i)|} \sum_{p\in P(i)} \log \frac{\exp(\Phi (r_i)\cdot \Phi (r_p)/\tau)}{\sum_{k \in A(i)} \exp(\Phi (r_i)\cdot \Phi (r_k)/\tau)}
\end{equation}
Here, we add a hyperparameter $\lambda$ to control the balance between cross-entropy and supervised contrastive learning loss in equation~\ref{eq:proposed1}.

By introducing equation~\ref{eq:proposed1}, we can simultaneously calculate the clusters between different representations $r$ and build the boundaries among different clusters in latent space. Therefore, the vectors for the same label in latent space are pulled together. Meanwhile, the cross-entropy takes charge of classifying the labels from the clusters sorted by supervised contrastive learning.

For the supervised contrastive loss, set $I$ in $\mathcal{L}_{SCL}$ means the four types of labels: \emph{O, COMMA, PERIOD, QUESTION}. Therefore, representations of all four labels are computed where the label contexts for punctuation and non-punctuation are fully incorporated within equation~\ref{eq:proposed3}.

The $\Phi$ denotes the $\ell_2$ normalization of $R$ after we get the $R$ from $\text{Encoder}(X)$. Since we get $\ell_2$ normalization on $R$. Therefore, the dot product between $r_i$ and $r_p$ (or $r_i$ and $r_k$) means the cosine similarity between the representation pair since the normalized representations exclude the influence from their magnitudes. Meanwhile, adding $\ell_2$ on $r$ avoids the issue that calculates the $\exp$ on a large number, which might cause the infinity error in the empirical practice. The $N$ is the number for the samples that belong to the labels that differ from $i$. 

The $\tau$ is the temperature to control the smoothness of the whole calculation. Small $\tau$ is beneficial to the training process since it is sensitive to the change of similarity score, which causes fluctuation for a slight change between two $r$. However, this would be hard to train since the loss and gradient for small $\tau$ are numerically unstable, which results in under-fitting.

\subsubsection{Supervised contrastive learning on language model}
As shown in Figure~\ref{fig:our contrastive learning}, we adopt the above loss onto a pre-trained language model. The combined loss function calculates the gradients based on different aspects regarding the clusters of representations and the nature of the classification task. 

We trained the language model on the punctuation restoration with equation~\ref{eq:proposed1}. The hidden states from the last layer of the language model are used as the representations for the punctuating tasks since they should have captured the context information throughout the feed-forward pass from previous layers. Meanwhile, by fine-tuning the task, the hidden states' distribution from the last layer is closer to the distribution of ground truth than those from previous layers. Therefore, we conduct supervised contrastive learning on the hidden states from the language model's last layer, which is denoted as $R$. Clusters $r_i \in R$ from different labels are pushed away that create spaces for cross-entropy to build boundaries on the sparse latent space tractably.

\section{Experiments}\label{sec:exp}
\subsection{Dataset}
IWSLT TED Talk dataset ~\cite{dataset} is a commonly used dataset for punctuation restoration tasks. This dataset consists of $2.1M$ words in the training set, $295k$ words in the validation set from the IWSLT20212 machine translation track, and $12.6k$ words (reference transcripts) in the test set are from the IWSLT2011 ASR track. Each word is labeled by one of four classes: \emph{O, COMMA, PERIOD, QUESTION}. The ratios of these four labels are $85.7\%$, $7.53\%$, $6.3\%$, $0.47\%$ respectively. The detail of the data distribution can be found in Table~\ref{tab:data}. In the preprocessing phase, we convert all words in the dataset into lower case, which prevents the sentence segment information leakage from word capitalization.

\subsection{Metrics}
The prediction results are evaluated by precision ($P$), recall ($R$), and $F_1$ score ($F_1$). Since we only focus on punctuation restoration task performance, the correctly predicted no punctuation would be ignored. The metrics on \emph{COMMA, PERIOD, QUESTION} are evaluated on the test set, and the overall metrics on those three punctuation marks are calculated as well.

\begin{table*}[t]
\caption{Model comparisons on IWSLT2011 Ref dataset. $P$, $R$, $F_1$ denote the Precision, Recall, and $F_1$ Score on test dataset respectively}
\label{tab:model_comparison}
\centering
\begin{threeparttable}
\begin{tabular}{lllllllllllll}
\toprule
\multirow{2}{4em}{\textbf{Models}}& \multicolumn{3}{c}{{\emph\textbf{COMMA}}} & \multicolumn{3}{c}{{\emph\textbf{PERIOD}}} & \multicolumn{3}{c}{{\emph\textbf{QUESTION}}} & \multicolumn{3}{c}{{\emph\textbf{OVERALL}}} \\
\cmidrule{2-13}
                    & $P$   & $R$  & $F_1$& $P$    & $R$  & $F_1$& $P$      & $R$  & $F_1$& $P$     & $R$  & $F_1$\\
\midrule
T-BRNN-pre ~\cite{T-BRNN-pre}                                   &      65.5 &      47.1 &      54.8 &      73.3 &      72.5 &      72.9 &      70.7 &      63.0 &      66.7 &      70.0 &      59.7 &      64.4 \\
BLSTM-CRF ~\cite{crf_ensemble}                                  &      58.9 &      59.1 &      59.0 &      68.9 &      72.1 &      70.5 &      71.8 &      60.6 &      65.7 &      66.5 &      63.9 &      65.1 \\
Teacher-Ensemble ~\cite{crf_ensemble}                           &      66.2 &      59.9 &      62.9 &      75.1 &      73.7 &      74.4 &      72.3 &      63.8 &      67.8 &      71.2 &      65.8 &      68.4 \\
DRNN-LWMA-pre ~\cite{DRNN-LWMA-pre}                             &      62.9 &      60.8 &      61.9 &      77.3 &      73.7 &      75.5 &      69.6 &      69.6 &      69.6 &      69.9 &      67.2 &      68.6 \\
Self-attention ~\cite{TRANS_NOT_BERT}                           &      67.4 &      61.1 &      64.1 &      82.5 &      77.4 &      79.9 &      80.1 &      70.2 &      74.8 &      76.7 &      69.6 &      72.9 \\
Multi-task Learning ~\cite{2020_Disflu_BERT}                    &      68.2 &      68.8 &      68.5 &      81.2 &      81.3 &      81.2 &      81.2 &      81.3 &      82.1 &      76.8 &      77.1 &      77.2 \\
Bert-Punct-BASE ~\cite{BERT1}                                   &      72.1 &      72.4 &      72.3 &      82.6 &      83.5 &      83.1 &      77.4 & \bf{89.1} &      82.8 &      77.4 &      81.7 &      79.4 \\
Focal Loss ~\cite{focal_loss_bert}                              &      74.4 &      77.1 &      75.7 & \bf{87.9} &      88.2 &      88.1 &      74.2 &      88.5 &      80.7 &      78.8 &      84.6 &      81.6 \\
Focal Loss\tnote{*}                                             &      68.8 &      71.7 &      70.2 &      82.5 &      82.8 &      82.7 &      78.5 &      86.5 &      82.1 &      76.4 &      80.4 &      78.3 \\
Self-Ensemble (RoBERTa$_{\text{LARGE}}$) ~\cite{acl2020-punc}    &      74.3 & \bf{76.9} &      75.5 &      85.8 & \bf{91.6} & \bf{88.6} &      83.7 & \bf{89.1} &      86.3 &      81.3 & \bf{85.9} &      83.5 \\
Self-Ensemble (RoBERTa$_{\text{BASE}}$) ~\cite{acl2020-punc}     &      76.9 &      75.4 & \bf{76.2} &      86.1 &      89.3 &      87.7 &      88.9 &      87.0 &      87.9 &      84.0 &      83.9 & \bf{83.9} \\
Token-Level SCL (Ours)                                          & \bf{78.4} &      73.1 &      75.7 &      86.9 &      87.2 &      87.0 & \bf{89.1} & \bf{89.1} & \bf{89.1} & \bf{84.8} &      83.1 & \bf{83.9} \\
\bottomrule
\end{tabular}
\begin{tablenotes}\footnotesize
\item[*] This is the result of our implementation for focal loss
\end{tablenotes}
\end{threeparttable}
\vspace{-4mm}
\end{table*}


\begin{table}[t]
\caption{$F_1$ Scores on cross-entropy (CE) and our supervised contrastive learning (SCL)}
\label{tab:loss_comparison}
\centering
\begin{tabular}{cccc}
\toprule
Loss & $\text{BERT}_{\text{BASE}}$ & $\text{RoBERTa}_{BASE}$ & $\text{RoBERTa}_{LARGE}$ \\
\midrule
CE   & 76.4                                & 78.5                    & 80.9                     \\
SCL  & 79.6                                & 80.4                    & 83.9                     \\
\bottomrule
\end{tabular}
\vspace{-4mm}
\end{table}

\subsection{Experiment setup}
We examined our proposed methods onto two language models: RoBERTa ~\cite{RoBERTa} and BERT ~\cite{BERT}. We set the max length of the input token to $256$ with a batch size of $32$. Adamw optimizer ~\cite{adamw} is used in our experiment with a learning rate of \num{1e-5}. The dropout rate in our experiment is $0.1$. Figure~\ref{fig:our contrastive learning} shows the architecture of our model. We place only one linear layer after the language model to exclude the potential performance gain by other deep network structures from the contrastive loss. Meanwhile, fine-tuning the language model on merely one extra layer also shows that supervised contrastive learning separates the representations from different classes by making spaces among different clusters, creating a tractable space for one linear layer to build boundaries. We trained our model for $20$ epochs, and the validation set is used to select the best model. The evaluation results are performed on the test set. The training process jointly fine-tunes the parameters of the entire network.

\subsection{Result}
The best result reported in Table~\ref{tab:model_comparison} is based on $\text{RoBERTa}_{\text{LARGE}}$, which has $355M$ parameters, $16$ attention heads, and $24$ layers of transformer encoders. The $\lambda$ and temperature used in the best result are $0.1$ and $0.6$. To stabilize the training, we divide the temperature by a base temperature of $0.07$ to get a final temperature $\tau$, which means $\tau = temperature/base\ temperature$. 

Table~\ref{tab:model_comparison} shows that our model has the best result on \emph{QUESTION}, which is the smallest class. 
We attribute this to the alleviation of the data imbalance problem by our proposed method.

\subsection{Comparison with other models}
\emph{T-BRNN-pre, BLSTM-CRF, Teacher-Ensemble, DRNN-LWMA-pre, Bert-Punct BASE, Multi-task Learning, and Focal Loss} are trained with only text data, while the \emph{Self-attention} model utilized both lexical and prosody features. We use the lexical data to train our model. The models based on transformer architectures outperform those on RNNs. Overall, our model achieves the best result among all the baseline methods listed in Table~\ref{tab:model_comparison}. 

\subsection{Comparison with focal loss}
Since we have a similar base structure with the focal loss model~\cite{focal_loss_bert}, we implemented the focal loss on our experiment setting as one of the baseline methods. The model and focal loss settings used in the experiment are the same as the proposed paper, $\text{BERT}_\text{BASE}$ model with $\gamma=2$ for focal loss. Our result over $\text{BERT}_\text{BASE}$ model is shown in Table~\ref{tab:loss_comparison}. By adopting token-level supervised contrastive learning, we still gain $1.3\%$ definite improvement on the overall $F_1$.

\begin{table}[t]
\caption{Label distributions of the IWSLT dataset}
\label{tab:data}
\centering
\begin{tabular}{llll}
\toprule
         & Train     & Validation & Test   \\
\midrule
Empty    & 1,801,727 & 252,922    & 10,943 \\
Comma    & 158,392   & 22,451     & 830    \\
Period   & 132,393   & 18,910     & 807    \\
Question & 9,905     & 1,517      & 46     \\
\bottomrule
\end{tabular}
\vspace{-4mm}
\end{table}

\subsection{Comparison with cross-entropy}
To demonstrate the effectiveness of our method, we made comparisons between our method and cross-entropy over the same experiment settings in Table~\ref{tab:loss_comparison}. Here we can see over $1.9\%$ absolute $F_1$ improvement on all three language models. The $\text{BERT}_\text{BASE}$ benefits the most from our method with an improvement of up to $3.2\%$ absolute $F_1$ score. 

\subsection{Comparison with self-ensemble}
Our approach has comparable results to the self-ensemble method. The self-ensemble method uses RoBERTa as the base language model followed by the point-wise feed-forward network with the dimension of $1568$. Meanwhile, this method applies the mechanism of the sliding window to get multiple predictions for a token. Our method only has one layer after the RoBERTa$_{\text{LARGE}}$ without any ensemble mechanism. On the RoBERTa$_{\text{LARGE}}$ model, we achieved $83.9$, higher than the RoBERTa$_{\text{LARGE}}$ through the self-ensemble method, with fewer layers and direct output, showing
more efficiency than the self-ensemble method.


\section{Conclusion}\label{sec:conlusion}
This paper leveraged the token-level supervised contrastive learning to alleviate the data imbalance problem in punctuation restoration. The results on the IWSLT2011 dataset showed the ability of supervised contrastive learning with one linear layer after the transformer-based language model. Incorporating supervised contrastive learning into the task yielded absolute performance improvement on $F_1$ from $1.9\%$ to $3.2\%$. Also, our method achieved comparable results with the ensemble models. 
In future work, we plan to conduct automatic data augmentation and other techniques onto supervised contrastive learning that enables better performance and robustness in both supervised and semi-supervised learning settings.

\bibliographystyle{IEEEtran}

\bibliography{main}


\end{document}